\documentclass[anon,12pt]{colt2020} 

\title[]{Survey of Matrix Completion Algorithms}
\usepackage{times}
\usepackage[noend]{algorithm,algorithmic}
\usepackage{caption}
\usepackage{amsmath, amssymb}
\usepackage[utf8x]{inputenc} 
\usepackage{hyperref}
\usepackage{interval}
\usepackage{caption}
\usepackage{hyperref}
\usepackage{cleveref}%

\usepackage{natbib}
\usepackage{tikz}

\definecolor{darkgray}{rgb}{0.50, 0.50, 0.50}
\definecolor{gray}{rgb}{0.70, 0.70, 0.70}
\definecolor{lightgray}{rgb}{0.92, 0.92, 0.92}

\coltauthor{%
 \Name{Jafar Jafarov} \Email{jafarov@alumni.stanford.edu}\\
 \addr Stanford University%
}

\begin{document}

\hspace{15mm} \LARGE Survey of Matrix Completion Algorithms \\[2ex]

\large 

\textbf{Jafar Jafarov} \hspace{60mm} jafarov@alumni.stanford.edu

Stanford University

\vspace{3mm}
\begin{abstract}
Matrix completion problem has been investigated under many different conditions since Netflix announced the Netflix Prize problem. 
Many research work has been done in the field once it has been discovered that many real life dataset could be estimated with a low-rank matrix.
Since then compressed sensing, adaptive signal detection has gained the attention of many researchers.
In this survey paper we are going to visit some of the matrix completion methods, mainly in the direction of passive and adaptive directions.
First, we discuss passive matrix completion methods with convex optimization, and the second active matrix completion techniques with adaptive signal detection methods.
Traditionally many machine learning problems are solved in passive environment. 
However, later it has been observed that adaptive sensing algorithms many times performs more efficiently than former algorithms.
Hence algorithms in this setting has been extensively studied.
Therefore, we are going to present some of the latest adaptive matrix completion algorithms in this paper meanwhile providing passive methods.\\
\end{abstract}


\vspace{7mm}
\section{Introduction}
\vspace{3mm}

Netflix has announced a problem of estimating unavailable entries of partially filled matrix and since then approximating unknown entries have been focus of many researchers \cite{rennie2005fast, srebro2004learning}.
The problem was targeting to estimate how a user would rate a movie that he / she never watched based on previously provided ratings.
The problem has been motivated in an environment that, the company had the data of rating of some users on some subset of movies.
Having this accurate estimation method provided, it would be used for the movie recommendation of the system for the users.\\[2ex]
Surprisingly, it has been discovered that the underlying structure of user movie rating data has almost low rank structure.
It means that, if the entire matrix was available the column space of it would be low rank subspace with additional noise structure on it.
Hence, this column space could be formalized as the most columns those are belong to the highest values of singular values.
Since then, low-rank estimation has gained attention from the researchers to find an efficient way to estimate this structure \cite{woodruff2014sketching}.\\[2ex]
Matrix completion has been studied both in passive and active setting.
One of the earliest works in this space has been done in the space of collaborative filtering has been done in \cite{goldberg1992using}.
Later, many works has been done in the convex optimization and optimization relaxations space as well.
Many research papers has investigated this methodology by various mathematicians \cite{tao, recht1, recht2} and nuclear norm minimization set a standard how to approach to this kind of problems.\\[2ex]
In traditional machine learning algorithms, data is provided in the beginning of the algorithm, and the task of the algorithm is to study the data, make correct conclusions or compute correct parameters based on the existing data.
And apply the conclusions or computed parameters to the upcoming test data.
In adaptive learning methods, the learning process is ongoing dynamic process, and the set of data used for learning changes progressively. 
Many times flow of the data is decided by the learning algorithm itself, based on analysing currently available information set and querying new data based on that.\\[2ex]
It has been shown that adaptivity helps in many machine learning problems. In generic learning problem \cite{riedmiller1992rprop}, in distributed optimization problem \cite{xu2017adaptive}, statistical analysis \cite{jamieson2016power} are just some examples to these problems.\\[2ex]
The idea of adaptivity has been applied to matrix completion as well.
One of the earliest works in the field is due to \cite{akshay1} which showed adaptive sampling in each column can help to identify linearly independent column vectors. 
Later authors further improved the existing result and also other researchers have further studied in this domain \cite{ramazanliadaptive, akshay2} and works has been extended to noise tolerant version in \cite{nina, ramazanli2022adaptive}.

\section{Preliminaries}
\vspace{3mm}

In this section, we provide basic notations and definitions that has been used through this survey.
We start with giving notations that has been used here first.
Then, we define important parameters that has been used in matrix completion techniques. Such as coherence - which has been used by many researchers in the investigation of compressed sensing, singular values and sparsity number.
\vspace{1mm}
\paragraph{Notations} The $n_1 \times n_2$ sized matrix we target matrix we try to recover is denoted by M. 
We assume that, it has a low rank, and rank values is given by $r$.\\[2ex]
For a given matrix $X \in \mathbb{R}^{n_1 \times n_2}$, its singular value decomposition is represented by $X = U \Sigma V^*$, where $U$ and $V$ have orthonormal columns those are basis for the column and row space of the matrix, and $\Sigma$ is a diagonal matrix, which its entries $\sigma_i = \Sigma_{ii}$ are uniquely determned by the initial matrix $M$.
We call the entries of the diagonal matirx by singular value. \\[2ex]
$\| .  \|_*$ stands for the nuclear norm for a given matrix and its value is equal to the sum of singular values of the matrix.
\begin{align*}
    \| X \|_* = \sum \Sigma_{ii}
\end{align*}
For a given subspace $W \subset \mathbb{R}^{n_1}$, we represent the orthogonal projection operator to this subspace by $\mathcal{P}_W()$.
In a different setting when $\Omega \subset [n_1] \times [n_2]$ the projection operator associated by $\Omega$ is defined as 
\begin{align*}
    \mathcal{P}_{\Omega} (X)_{ij}  = 
    \begin{cases}
     X_{ij} \hspace{12mm} if \hspace{2mm} (i, j) \in \Omega \\
     0 \hspace{16mm} otherwise
    \end{cases}
\end{align*}
where $[n]$ stands for the set $[n] = \{1, 2, \ldots, n \}$.\\[2ex]
Next important definition here is the coherence parameter which is defined as below.
Coherence of an $r$-dimensional subspace $W \subset \mathbb{R}^n$ is defined as
\begin{align*}
\mu(W) = \frac{n}{r} \underset{1 \leq j \leq n}{\max} || \mathcal{P}_{\mathbb{W}} e_j ||^2,
\end{align*}
and finally as mentioned in \cite{poczos2020optimal} the sparsity number is defined as 
\begin{align*}
  \psi(W)=\mathrm{min}\{\|x\|_0 | x\in W \text{ and } x\neq 0 \}  
\end{align*}
and sparsity-number of a matrix is just simply sparsity-number of its column space.
Similarly, we define the \textit{nonsparsity-number} of the subspace $W \subset \mathbb{R}^m$   as: 
\begin{align*}
 \overline{\psi}(W)= m - \psi(W)
\end{align*}
Authors of the paper, showed that idea of \text{sparsity-number} could be used in exact matrix completion in various settings \cite{ilqarsingle, ramazanli2022matrix}.

\vspace{5mm}
\section{Review of Algorithms}
\vspace{2mm}

In this section we  provide various matrix completion algorithms under different settings.
In the first subsection we talk about matrix completion in passive setting.
Second, we will talk about low-rank estimation with leverage scores.
Third we provide matrix completion algorithms with sparsity-number.

\vspace{3mm}
\subsection{Passive Matrix Completion Algorithms}
\vspace{1mm}

We start to this section with defining matrix completion problem in passive setting.
In this setting there exist just one stage of the available information set. Which is partially observed entries of the matrix. 
This entries are not necessarily to be placed in same column or row.
Moreover, we target to estimate / recover remaining entries using these observations.

\vspace{2mm}

\[ 
 \begin{bmatrix}
    \bf{1.3} & ? & ? & ? & \bf{2.1} & ? \\
    ? & \bf{3.2} & ? & ? & ? & \bf{-0.3} \\
    ? & ? & ? & ? & \bf{0.87} & ? \\
    ? & ? & ? & \bf{23} & ? & ? \\
    ? & \bf{0.3} & \bf{0.1} & ? & \bf{0.2} & ? \\
    ? & ? & ? & ? & \bf{0.1} & \bf{0.2} 
\end{bmatrix} 
\hspace{6mm} \rightarrow  \hspace{6mm}
\begin{bmatrix}
\bf{1.3} & 1.1      & 0.2      & 1.34    & \bf{2.1}  & -1.3 \\
3.1      & \bf{3.2} & 3.13     & 5.4     & 1.7       & \bf{-0.3} \\
3.3      & 1.1      & 2.67     & -0.34   & \bf{0.87} & 2.1 \\
2.1      & 1.2      & 1.12     & \bf{23} & 11.       & 1.76 \\
1.76     & \bf{0.3} & \bf{0.1} & -0.63   & \bf{0.2}  & -3.1 \\
-1.2     & 0.64     & 24.1     & 7.63    & \bf{0.1}  & \bf{0.2} 
\end{bmatrix} 
\]
\vspace{4mm}

\hspace{-6mm} Basically at the first stage we pick some of the entries of the matrix to observe.
One famous method how to pick these entries is just uniform sampling of entire matrix to pick the entries.
Then using the observations from the selected entries we recover unobserved entries.\\[2ex]
An interesting observation here is that the nuclear norm minimization is able to successfully to solve this problem.
\cite{recht1} have shown that the following convex optimization program returns the exactly recovered low-rank matrix:
\vspace{2mm}
\begin{align*}
    \text{minimize} \hspace{5mm}  &\| X \|_* \\
    \text{subject to} \hspace{5mm} &\mathcal{P}_{\Omega}(X) = \mathcal{P}_{\Omega}(M)
\end{align*}
\vspace{2mm}
and they have proved that the following theorem get satisfied.

\begin{theorem} \cite{recht1}
Let M be an $n_1 \times n_2$ sized matrix of rank r.
Then the convex optimization problem that is defined above could successfully recover the underlying matrix, whenever $| \Omega | > C n^{6/5} r \log{n}$ \hspace{0.5mm} where \hspace{0.5mm} $n = \mathrm{max}(n_1, n_2)$
\end{theorem}
\vspace{1mm}
Later, this result has been further improved to a tighter bound in a given condition.
In particular, in the paper \cite{recht2} the following theorem has been proven:
\vspace{1mm}
\begin{theorem} \cite{recht2}
Let $M$ be an $n_1 \times n_2$ sized matrix of rank $r$.
Then, the nuclear norm minimization suggested in (1) would recover the underlying matrix using $\| \Omega \| > 32 \mathrm{max}(\mu_0^2, \mu_1)$ entries with probability $1 - 6 \log{n_2}(n_1+n_2)^{2-2\beta} - n^{2-2\beta^{1/2} }_2$ whenever the following two conditions got satisfied:
\begin{itemize}
    \item The row and column space coherence number is bounded above by some positive $\mu_0$
    \item The matrix $UV^*$ has a maximum entry that is bounded above by $\mu_1 \sqrt{r/(n_1 n_2)}$
\end{itemize}
\end{theorem}
\vspace{3mm}
Later \cite{tao} proved a lower bound for the methods that is involving nuclear norm minimization, and result could be presented as below:
\begin{theorem} \cite{tao}
Suppose that M is an $n_1 \times n_2$ sized matrix with rank $r$, and $\mu_0 $ is larger or equal than one. 
Then under the uniform sampling observation process, there are infinitely with the fixed entry values given that \hspace{1mm} $ 0 < \delta < 1/2 $ \hspace{1mm} and 
\begin{align*}
m < (1-\epsilon) \mu_0 \mathrm{max}(n_1,n_2) r \log{ \frac{n}{2\delta} }  
\end{align*}
 where
\begin{align*}
 \epsilon = \frac{1}{2} \frac{\mu_0 r}{n} \log{\frac{n}{2\delta}}    
\end{align*}

\hspace{-6mm}Then there are infinitely many matrices that solves the convex optimization problem proposed above.
\end{theorem}

\vspace{4mm}

\subsection{Active Matrix Completion Algorithms}
\vspace{4mm}

In this section we will discuss about active matrix completion algorithms.
The main difference of the algorithms in this section compared to previous ones is that there are multiple stages of information set here.\\[2ex]
This means that in the first stage of the algorithm we don't have any information available.
Then we send query to observe some entries of the underlying matrix.
Based on the observed entries, we decide which entries would be the best to observe in the next stage.
This process keep going this way, and at each step we observe in a more informative way.
An example of several steps of the algorithm could be represented as below:\\[2ex]
\[ 
 \begin{bmatrix}
    ? & ? & ? & ? & ? & ? \\
    \bf{3.4} & ? & ? & ? & ? & ? \\
    ? & ? & ? & ? & ? & ? \\
    ? & ? & ? & ? & ? & ? \\
    \bf{-1.2} & ? & ? & ? & ? & ? \\
    ? & ? & ? & ? & ? & ? 
\end{bmatrix} 
\hspace{12mm} \rightarrow  \hspace{6mm}
 \begin{bmatrix}
    \bf{-1.1} & ? & ? & ? & ? & ? \\
    \bf{3.4} & ? & ? & ? & ? & ? \\
    \bf{-0.8} & ? & ? & ? & ? & ? \\
    \bf{0.4} & ? & ? & ? & ? & ? \\
    \bf{-1.2} & ? & ? & ? & ? & ? \\
    \bf{0.11} & ? & ? & ? & ? & ? 
\end{bmatrix} 
\hspace{4mm} \rightarrow
\]

\vspace{5mm}

\[ 
 \begin{bmatrix}
    \bf{-1.1} & \bf{1.13} & ? & ? & ? & ? \\
    \bf{3.4} & ? & ? & ? & ? & ? \\
    \bf{-0.8} & ? & ? & ? & ? & ? \\
    \bf{0.4} & \bf{2.1} & ? & ? & ? & ? \\
    \bf{-1.2} & ? & ? & ? & ? & ? \\
    \bf{0.11} & ? & ? & ? & ? & ? 
\end{bmatrix} 
\hspace{3mm} \rightarrow  \hspace{6mm}
 \begin{bmatrix}
    \bf{-1.1} & \bf{1.13} & ? & ? & ? & ? \\
    \bf{3.4} & 1.1 & ? & ? & ? & ? \\
    \bf{-0.8} & 1.4 & ? & ? & ? & ? \\
    \bf{0.4} & \bf{2.1} & ? & ? & ? & ? \\
    \bf{-1.2} & 0.8 & ? & ? & ? & ? \\
    \bf{0.11} & 0.45 & ? & ? & ? & ? 
\end{bmatrix} 
\]

\vspace{5mm}
\hspace{-6mm}In the example above, at each step we observe some entries in a given column. 
Then, based on our observation, we decide either to observe all other entries in this column or recover remaining entries based on these observed elements.
These type of algorithms has been studied by many researchers as in the following framework:
\vspace{5mm}

\begin{algorithmic}[1]
    \FOR{$i$ from $1$ to $n$}        
    \STATE Sample uniformly $\Omega \subset [m]$
    \STATE  \hspace{0.2in} \textbf{if} $\| \mathbf{M}_{\Omega:i}-{\mathcal{P}_\mathbf{\widehat{U}_{\Omega}^k}} \mathbf{M}_{\Omega:i}\| >0$ 
    \STATE  \hspace{0.4in}  $\widehat{\mathbf{U}}^{k+1} \leftarrow \widehat{\mathbf{U}}^{k} \cup \mathbf{M_{:i}} $
    \STATE  \hspace{0.4in} Orthogonalize $\widehat{\mathbf{U}}^{k+1}$
    \STATE \hspace{0.4in} $k=k+1$          
   \STATE   \hspace{0.2in}  \textbf{otherwise:} 
   \STATE \hspace{0.4in} $\widehat{\mathbf{M}}_{:i} = \widehat{\mathbf{U}}^k {\widehat{\mathbf{U}}^{k^+}_{\Omega}}
 \widehat{\mathbf{M}}_{\Omega :i}$
\ENDFOR
\end{algorithmic}
\vspace{4mm}
The first result we present in this framework is due to \cite{akshay1}. 
Authors have proved the following theorem using the ideas how to detect linearly independent column provided in the paper \cite{balzano2010high}.
\begin{theorem} 
Let M be an underlying $n_1 \times n_2$ sized matrix of rank$-r$ and column space coherence $\mu_0$. 
Assuming that $\Omega$ provided above in the algorithm satisfies the condition that its size is lower bounded by $\| \Omega \| > 36 r^{3/2} \mu_0 \log{ \frac{r}{\epsilon}}$  then the algorithm successfully exactly recovers the underlying matrix $M$ with probability of \hspace{1mm} $1-\epsilon$. 
\end{theorem}
\vspace{2mm}
The idea of the proof is to rely on concentration inequalities proved in  \cite{balzano2010high} to give condition to detect linear independence.  
This algorithm has been further studied by authors and result has been improved in a followup work from the authors of the aforementioned paper. 
In particular, in the following paper \cite{akshay2} authors show that, it is enough to pick $\Omega$ in a way that its size to be lower bounded by $32 r \log{r/\epsilon}$.
\vspace{2mm}
\begin{theorem} 
Let M be an underlying $n_1 \times n_2$ sized matrix of rank$-r$ and column space coherence $\mu_0$. 
Assuming that $\Omega$ provided above in the algorithm satisfies the condition that its size is lower bounded by $\| \Omega \| > 32 r \mu_0 \log^2{ \frac{r}{\epsilon}}$  then the algorithm successfully exactly recovers the underlying matrix $M$ with probability of \hspace{1mm} $1-\epsilon$. 
\end{theorem}
 \vspace{1mm}
Interestingly, it has been later shown that, the provided lower bound here still is not tight, and there is still room to further improvement.
Therefore, in the paper \cite{nina} have shown that the bound given here could be further optimized, such that selecting $\| \Omega \|$ just to be lower bounded by $8 r \mu_0 \log{ \frac{r}{\epsilon}}$ is enough for the algorithm.

\begin{theorem} 
Let M be an underlying $n_1 \times n_2$ sized matrix of rank$-r$ and column space coherence $\mu_0$. 
Assuming that $\Omega$ provided above in the algorithm satisfies the condition that its size is lower bounded by $\| \Omega \| > 8 r \mu_0 \log{ \frac{r}{\epsilon}}$  then the algorithm successfully exactly recovers the underlying matrix $M$ with probability of \hspace{1mm} $1-\epsilon$. 
\end{theorem}
\vspace{1mm}
Further optimization to this class of the algorithm has been done due to \cite{ilqarsingle}.
Authors used different parametrization for this class of algorithms to enrich the algorithm to perform near theoretical lower bounds.
\vspace{1mm}
\begin{theorem}
Let $M$ be a $n_1 \times n_2$ sized matrix of rank $r$, which has column space sparsity-number donated by $\psi_0$ and row space sparsity number donated by $\psi_1$.
Then the framework above recovers the underlying matrix $M$ whenever $|\Omega|$ is bounded above by 
$$\mathrm{min}\Big(  2 \frac{n_1 }{{\psi_0}}\log{(\frac{r}{\epsilon})} , \frac{\frac{2n_1}{\psi(\mathbb{U})}(r+2 +\log{\frac{1}{\epsilon}})}{\psi_1} \Big). $$
\end{theorem}
\vspace{2mm}
The authors has shown that this bounds is as small as $\mathcal{O}((n_1+n_2-r)r)$ in many cases which is equal to the degree of freedom of $n_1\times n_2$ sized rank $r$ matrices.\\[2ex]
Another class of active completion algorithms performs in a less structured way. 
In the previous set of algorithms we discussed, for any step we have only observed entries from the same column.
In the following algorithm, authors have presented a more interactive algorithm, which for every phase one new element get observed in each column.
The algorithm goes many phases like this until it detects $r-$many linearly independent column / row vectros.
\vspace{2mm}
\begin{algorithmic}[1]
    \WHILE{$\widehat{r} < r$}
    \FOR{$j$ from $1$ to $n$}
    \STATE Uniformly pick an unobserved entry $i$ from $\mathbf{M}_{:j}$  
    \STATE $\widehat{R}={R} \cup \{i\} , \widehat{C}= {C} \cup \{j\}$  
    \STATE \textbf{If} $ \mathbf{M}_{\widehat{R}:\widehat{C}}$ is nonsingular
    \STATE \hspace{0.1in} Fully observe $\mathbf{M}_{:j}$ and $\mathbf{M}_{i:} $ and set
     $R =\widehat{R}$ , $C = \widehat{C}$ , $\widehat{r}=\widehat{r}+1$
\ENDFOR
\ENDWHILE
\STATE Orthogonalize column vectors in $C$ and assign to $\widehat{\mathbf{U}}$
\FOR{each column $j \in [n]\setminus C$}
\STATE $\widehat{\mathbf{M}}_{:j} = \widehat{\mathbf{U}} {\widehat{\mathbf{U}}_{R:}}^+
 \widehat{\mathbf{M}}_{R: j}$
\ENDFOR
\end{algorithmic}
\vspace{7mm}
Authors of the paper used a parametrization called sparsity-number to make the analysis of the algorithm simpler.
The idea is to use the number of maximum zero coordinates of a nonzero element in the column space, to detect linear independence much easier compared to coherence parameter.
Hence, the following theorem has been proven in the paper:
\vspace{2mm}
\begin{theorem}\label{thm:lg2} 
Let M be an underlying $n_1 \times n_2$ sized matrix of rank$-r$ and column space coherence $\mu_0$. 
Assuming that $\Omega$ provided above in the algorithm satisfies the condition that its size is lower bounded by $\| \Omega \| > 8 r \mu_0 \log{ \frac{r}{\epsilon}}$  then the algorithm successfully exactly recovers the underlying matrix $M$ with probability of \hspace{1mm} $1-\epsilon$ using at most 
$$(n_1+n_2-r)r +  \mathrm{min}\Big(  2 \frac{n_1 n_2 }{{\psi_0}}\log{(\frac{r}{\epsilon})} , \frac{\frac{2n_1}{\psi_0}(r+2 +\log{\frac{1}{\epsilon}})}{\psi_1}n_2 \Big). $$
many observations
\end{theorem}
\vspace{2mm}
Using the idea of sparsity-number, authors have solved the matrix completion under various pre-information setting.
In one of the algorithms authors just needed to have the assumption of either row or column space not being very coherent which is minimal amount of information.
The works has been extended to heterogeneous cost completion problem which different entries of the matrix has different cost, and target to recover the underlying matrix as cheap as possible.\\[2ex]
Adaptive matrix completion with respect to query complexity has been extensively studied.
One of the earlier active completion algorithm is due to the paper \cite{chakraborty2013active}. 
Which later this question has been further explored in the paper \cite{ruchansky2015matrix}.\\[2ex]
Adaptive matrix completion has been studied for the matrices in special structure as well \cite{bhargava2017active}.
Later matrix completion algorithms in the special structured matrices has been explored in \cite{ruchansky2016matrix}.

\bibliography{ms}



\end{document}